# Writing user personas with Large Language Models

# Testing phase 6 of a Thematic Analysis of semi-structured interviews

**Stefano De Paoli**

**Abertay University, Sociology Division**

**s.depaoli@abertay.ac.uk**

**Abstract**

The goal of this paper is establishing if we can satisfactorily perform a Thematic Analysis (TA) of semi-structured interviews using a Large Language Model (more precisely GPT3.5-Turbo). Building on previous work by the author, which established an embryonal process for conducting a TA with the model, this paper will perform a further analysis and then cover the last phase of a TA (phase 6), which entails the writing up of the result. This phase was not covered by the previous work. In particular, the focus will be on using the results of a TA done with the LLM on a dataset of user interviews, for writing user personas, with the model building on the TA to produce the personas narratives. User personas are models of real users, usually built from a data analysis like interviews with a sample of users. User personas are tools often used in User Centered Design processes. The paper shows that the model can build basic user personas with an acceptable quality deriving them from themes, and that the model can serve for the generation of ideas for user personas.

## Introduction

The problem of this paper is to understand if we can use a Large Language Model (specifically GPT3.5-Turbo) to perform a Thematic Analysis (TA) of semi-structured interviews, also focusing on the last phase of a TA, which entails the writing up of the results. This work is conducted within a process of Human-AI collaboration, a concept in the field of Artificial Intelligence (AI) that assumes that humans and AI systems can collaborate to achieve goals and tasks (see e.g. Vössing et al., 2022, Siemon, 2021 for a general overview; and Jiang et al. 2021, specifically for qualitative analysis reflections on this). Elsewhere, I presented an embryonal, and perhaps still crude, process for conducting a TA with the same LLM, following some of the key phases to a TA proposed by Braun & Clarke (2006) in their seminal work. These 6 phases include the human analyst(s): familiarisation with the data (phase 1), generation of initial codes, i.e. relevant features in the data (phase 2, done inductively in my case), the generation of themes, or patterns in the data, based on sorting the code in themes (phase 3), confirming the validity of themes, by reviewing them (phase 4), the renaming and summarising of themes, to confirm their validity (phase

5) and the writing up of the results, as an integral part of TA (phase 6). In my previous work I argued that the LLM can only reasonably perform phases 2-5 of a TA. Phase 1 and phase 6 were not tackled directly, either because the memory and tokens limits of the LLMs do not allow the LLM to perform the work (phase 1 – familiarisation) or because of potential ethical issues (phase 6 – writing the report). In this paper, I would like to tackle some aspects associated with phase 6 of a TA and the process of writing up the results.

The focus of this paper is methodological: establishing the validity of the use of LLMs for the conduction of qualitative analysis of data (specifically TA), with a focus on semi-structured interviews and largely in the context of social sciences. As I discussed in my previous paper, one of the epistemological challenges for this is that qualitative analysis is normally done by humans through interpretation of meaning, and this is something that LLMs are not necessarily capable as they operate on the language from a computational, logical and structural perspective (Floridi, 2023). Nonetheless, it has been possible to show that an LLM can perform something looking like a basic inductive TA with at least some degree of validity, through a qualitative comparison with the work of human analysts (De Paoli, 2023). Other authors have used Cohen's kappa to confirm inter-reliability between an LLM and human conders, based on deductive coding processes (Xiao et al., 2023; Gao et al., 2023). TA can indeed be done with a deductive approach (where the grid of analysis is decided before-hand) or inductive where codes and themes are generated bottom-up from the data. I am interested in performing with LLMs an inductive approach to TA. It remains therefore to be assessed whether we can also cover phase 6 of the TA. For Clarke & Braun (2013): "Writing is an integral element of the analytic process in TA (and most qualitative research). Writing-up involves weaving together the analytic narrative and (vivid) data extracts to tell the reader a coherent and persuasive story about the data and contextualising it in relation to existing literature.". As such writing cannot be detached from the process of doing a TA. Therefore, to assess if an LLM can satisfactorily conduct a TA, we need to attempt also phase 6 with the model.

There is discussion about using LLMs to write research work, and there clearly are important ethical implications associated with this (see Lund et al. 2023 for an overview), whilst we have also seen authors citing these models as co-authors (e.g. King & chatGPT). On the other hand, journals and scholarly publishers have begun to create policies to clarify what is the acceptable use of these models in academic publishing. These, correctly in my view, start to be very stringent. For example, the editor-in-chief of the journal Science stated that they decided to update their "license and Editorial Policies to specify that text generated by ChatGPT (or any other AI tools) cannot be used in the work, nor can figures, images, or graphics be the products of such tools. And an AI program cannot be an author." (Thorp, 2023, p. 313). There also have been cases where publishers have removed chatGPT from the list of authors of already published papers (e.g. O'Connor, 2022).

We can agree with Thorp when he points out that AI manipulated images or text which is not produced by an author should not be included in academic publications and may largely amount to academic misconduct. However, we should consider that using LLMs to write research results may also entail working on intermediate phases of the writing process and within a Human-AI collaboration approach, or on scientific products which may have other applied use for other research activities. For Braun & Clarke (2006), phase 6 of TA, is an integral part of the analysis method, and therefore we should explore if also this phase can be performed with the support of an LLM. This paper will attempt at tackling this problem.

In this way the focus of the manuscript is not on the ethics discourses around the use of LLMs for writing scholarly work, but it is rather methodological. I will not propose a process to use the LLM to write a full paper (or part of it), rather I will focus on using the LLM to write intermediate narratives/models, which can be used to support other research activities. I will concentrate on using the LLM to write user personas, based on the results of a TA (of semi-structured interviews) conducted by the LLM in cooperation with the author.

User personas are an ideal candidate for the goal of this work. The Norman&Nielsen Group (NNG) defines a user persona as "a fictional, yet realistic, description of a typical or target user of the product. It is used to promote empathy, increase awareness and memorability of target users, prioritize features, and inform design decisions." (Harley, 2015). User personas are an important and established tool in user-centered design (UCD) and contribute significantly toward the creation of e.g. tools or services and help the designers to focus on people and specific aspects of them, during the design activities. A persona is normally a narrative with a variety of additional details (such as a picture, or specific traits of the user). They are built with a focus on identifying key aspects of the target users for a new design (or re-design), which include for instance the identification of the user needs, of their pain points, objectives, attitudes, behaviours, skills and so on, all of which should contribute to inform design decisions about e.g. a digital service. A useful overview of the advantages of user personas is provided by Miaskiewicz & Kozar (2011), who reviewed how several key authors have formulated these advantages. For a better understanding of the perception and importance of personas in UCD readers can consult for example the papers by Nielsen & Storgaard (2014) or Matthews & Whittaker (2012).

There may also be different types of personas used in UCD, which can support the design work in different ways. Again NNG, for example write that there can be broad and narrow personas (Salazar, 2020). The first is based on shallow data and can serve high level decisions. The narrow ones instead are based on more granular data and can support more specific decisions. Likewise, NNG stipulates that personas can be lightweight (or proto persona), qualitative and statistical (Laubheimer, 2020). The last two in particular are based on solid empirical evidence, with the statistical personas based on large quantity of data (but very costly to produce) about the target uses and the qualitative personas (which are the most widely used in UCD) that are based on the use qualitative research data, like interviews conducted with a sample of target users or with ethnographic observations of e.g. the target users at work. It is the qualitative and narrow personas I would like to focus on. For building qualitative personas a common way to gather data is with semi-structured interviews with a sample of the potential target users, and then an analysis is done on the interviews to identify patterns across the interviews. These patterns are the basis for building the personas narrative, including e.g. their background, goals, preferences or challenges. This is because recurring patterns signal potential common user needs or pain points which can and should then be the focus of the UCD work. TA is one of the analysis methods that can be adopted for the identification of these patterns across the interviews (see for example Turner et al., 2013 or Rosala, 2019).

We can therefore understand that (qualitative) user personas are an intermediate product of the user research in UCD, which can support the work of designers, and that they are the results of collecting empirical data and of qualitative analysis. If, as I suggested previously, we assume that that an LLM can perform at least in an embryonal form an inductive TA of semi-structured interviews (Phases 2-5 according to Braun & Clarke), we can explore whether the LLM can satisfactorily produce personas narratives based on this same analysis, or at least produce something which has some semblance with a user persona. Personas can therefore potentially be a good example of textual result here we can attempt to cover phase 6 of a TA as proposed by Braun & Clarke. This will be to focus of the following pages.

## Literature review

There is limited research on the use of LLMs for qualitative analysis. For example, a working paper by Mesec (2023) provided a comparative study between chatGPT and a human analyst based on one interview. However, this approach does appear rather simplistic as it just leverages the web chat application (https://chat.openai.com/), rather than use the model via an API to process data with some scale. This mirrors some tutorials that are available online[1], which propose basic elements of qualitative analysis, leveraging the web chat version of the model, and needing to manually paste interview transcripts in chatGPT and copying the results in e.g. text processing files. More solid existing research largely focusses on the process coding (covering mostly the equivalent of phase 2 of a TA), and as far as the author's understanding these works adopted a deductive approach to analysis (Xiao et al., 2023; Gao et al., 2023). The approach I am exploring is instead entirely inductive where the model operates without a prior coding framework. Moreover, in my opinion these works (nor indeed the online tutorials) do not align or propose any significant methodological discussion or reflection, and they seem to assume we can do a qualitative analysis with an LLM, with limited engagement of what this means for the methodological rigour of social research. They do not much reflect on what are the implications of doing social sciences qualitative analysis with the support of an LLM, or if indeed we should attempt qualitative analysis with an LLM. My approach is more cautious, and I propose that methodological issues should be put into focus alongside our attempts to establish whether we can indeed perform a qualitative analysis with LLMs. For example, we still do not know if academic journals will accept a paper which adopts an LLM performed TA, and we still need to find agreement if this has sufficient scientific rigour. Nonetheless, in particular the work by Xiao et al. (2023) and Gao et al. (2023) are important because they raise attention to several important issues, which include the crucial role of the prompt to the model (see later), the human-AI collaboration in qualitative analysis with LLMs, the use of LLMs for doing intermediate steps of a full analysis (like e.g. initial coding, or team discussion). The use of LLMs for qualitative analysis does seem indeed a potentially nascent area of inquiry, and therefore there is limited literature, but also a need to develop processes which are methodologically appropriate and accepted by the scientific community.

Some words need also to be spent in relation to personas, and specifically in relation to the use of LLMs for the generation of personas. As LLMs can easily produce textual outputs from a prompt, we have seen propositions to use chatGPT (again the web chat version) for producing user personas. There is material such as online videos or blog posts which explain how to generate a persona with chatGPT, with prompts such as: "Generate a user persona about a busy mum who lives in a city that wants to make sure her kids eat healthy without compromising time away from her hobbies and career"[2]. Prompt such as these can be used within the web chat to generate a user persona, with some believable details including a narrative and demographics. Further, it is possible to ask the model to write the output received "in the format of a user persona" and the model will propose lists of e.g. pain points, needs attitudes and so on. As it has been noted, there is something powerful in the model since it knows what a user persona is[3] without needing any contextual explanation. This is a relevant observation as later in the analysis and the generation of personas, we will assume the LLM knows what a user persona is. However, these personas generated with the LLM without being based on any empirical material are entirely fictional, and do not have the realistic component that the definition of a persona seen earlier implies ('fictional, yet realistic'). Whilst they look believable, they seem unlikely to be a good and realistic representation of the potential target users for UCD. Kocaballi (2023) has proposed a more structured approach, rather than just simply asking the LLM to generate a persona, to emulate an entire UCD process with an LLM, including the generation of personas. However, even if more structured, this work is entirely fictionally created by the model. So far so that, that there have been invitations to avoid using LLMs for generating personas[4], since these cannot be a true representation of target users for a product or service, and LLMs cannot replace the work of user research which requires inquiring with real users (e.g. with interviews or ethnographies). This debate is at the moment taking place largely on websites and other online fora, rather than on academic publications, and it is symptomatic of larger processes of automation of human work and of the oversimplifications which are possible with LLMs. It is not my intent to enter this specific debate, especially in the field of user research.

---

[1] See e.g. https://www.youtube.com/watch?v=9yN2bmf6BNE or https://www.youtube.com/watch?v=8dTs7D42ge0
[2] See e.g. https://www.youtube.com/watch?v=watkVh1U_ko

However, I believe it is rather different to use the LLM to generate user personas fictionally, without any underlying empirical research, and have instead personas generated based on some form of TA and on real qualitative data, within a Human-AI research collaboration mode. If the LLM (with the support of the human researcher) can produce at least satisfactorily some forms (or at least ideas) of user personas based on a data analysis, we may also be able to make a step forward toward covering phase 6 of a TA.

## Research design

This paper proposes a research design which mirrors the one I proposed in my previous paper (De Paoli, 2023). I will, in particular, cover phases 2-4 of a TA which will include an initial generation of codes, the grouping of the codes in themes, and a validation of themes. Phase 5, which covers the renaming of themes, will not be covered here (largely for keeping the text compact), but it certainly is possible also to operate this phase if needed. Readers can consult the specificities of the proposed (albeit embryonal) approach to a TA with an LLM in De Paoli (2023). Here I will concentrate on a shorter summary of the method-process for performing a TA, and then focus on covering phase 6. In the previous publication I also reflected on the limits of the LLM I am using (GPT3.5-Turbo), including on the limits to the number of tokens and the lack of memory, which impact the processing of large amounts of text. Again, the reader can see these reflections in the previous paper.

First, we need to provide some definitions. We have already mentioned earlier the 'prompt', this is essentially what we ask the model to produce. The output/response is what the model produces based on its understanding of the prompt instructions. Figure 1, shows a simple prompt and the response, as generated by the chat version of GPT.

---

[3] See https://www.persona-institut.de/en/chat-gpt-3-und-personas/
[4] See e.g. https://uxdesign.cc/please-do-not-use-chatgpt-to-generate-personas-85ffeaa6690b

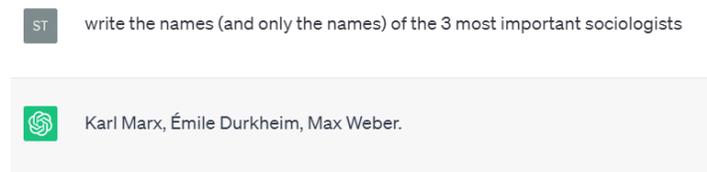

**Figure 1** – Prompt and response of chatGPT (web chat)

Second, to perform this research I used the model via the OpenAI API (https://openai.com/blog/openai-api), and not the web chat. The API allows to call the model inside python scripts. The advantage of this (over the use of the chat version) is that we can work with data and do manipulation and parsing of the data. It is because of this that we can have the model process real textual data with some degree of scale, upon which attempting to perform a data analysis (like a TA). In this paper the specificities of the python scripts will not be discussed unless necessary, whilst more space will be given to a discussion of the prompts.

As my goal is to build personas from a TA of semi-structured interviews, I selected an open dataset (covered by a Creative Commons license), available from zenodo. This ensures previous anonymisation of the data and the possibility to reuse this data. The selected dataset is the farmers' interviews (n=14) conducted by the project EUREKA, which is a subset of a larger dataset (Vágó, & Spanoghe, 2023). EUREKA was an H2020 project which aimed "to strengthen and improve the flow of agricultural and rural-related knowledge and innovation at European, national and regional level."[5]. EUREKA has produced a "first working version of the so-called EU FarmBook" digital platform. Interviews were conducted with several user groups including farmers, forestry workers or various advisors, in different European broad regions (including e.g. the Mediterranean or the North Sea regions). These interviews were conducted to produce user personas and user journeys, and therefore the research was planned with a focus on UCD. For the farmers' group, six personas were prepared and are available on the project website[6]. Therefore, we can consider this a potentially good dataset for the task of using an LLM to perform the writing up of a TA in the form of user personas.

Due to the maximum limit of the tokens in the model (4097), interviews had to be broken down into smaller chunks. From the initial 14 interviews, 31 chunks were created ranging between 700 to 1600 tokens (except for one short interview, which was already around 400 tokens). Data cleaning and preparation was necessary before the analysis, like removing special characters (e.g. text in Cyrillic for example, which had an English translation in the interview transcript) and saving all the interviews in plain text files.

The design I propose here is based on performing three steps: (1) first perform a TA with LLM on the interview chunks, replicating and enriching the process already proposed in my previous work; (2) using the LLM to create personas based on the results of the TA, with appropriate prompting; (3) offer an evaluation of the results, by looking at which themes and codes were used by the LLM to create one persona.

---
[5] See https://h2020eureka.eu/da/about
[6] See https://h2020eureka.eu/personas-and-user-journeys

## Performing the TA with the LLM

### Generating codes (phase 2)

Phase 2 of a TA involves generating inductively the codes from the data. Since the goal here is to try to develop user personas, I asked the model to identify separately codes for the users' a) needs and b) challenges. For this I used the prompt presented in Figure 2. This prompt is similar to the one I proposed in my previous work, however I added to the prompt a definition of what is a challenge. This definition in part was taken from a report of the EUREKA project (Van der Cruyssen, 2021), where a challenge was defined in relation to "access to data and knowledge" and, to specify this better, I added a simpler definition of a challenge. For the separate generation of user needs, I used in the prompt a definition from the UK Government (2017) which I consider simple and clear which defines user needs as "the needs that a user has of a service, and which that service must satisfy for the user to get the right outcome for them". Considering the limits of the max tokens of the model, I asked the model to generate up to 2 challenges and up to 3 needs from each interview chunk. Each chunk of interview is passed into the prompt via a for-loop in the variable 'text'.

```
prompt = f"""
    A challenge is an obstacle that a person tries to overcome, in particular in relation to accessing data and knowledge.

    Identify up tp 2 relevant challenges for the interviewee in the text below, provide a name for each challenge,
    a summary description of the challenge and a quote from the respondent for each interest no longer than 4 lines

    Format the response as a json file keeping names, descriptions and quotes togeter in the json, and keep them
    together in 'Challenges'.

    ```{text}```
    """
```

**Figure 2** – Prompt for initial coding of challenges

This prompt allowed the identification of n=62 user code-challenges and n=93 user code-needs from the interviews. Due to possible duplications (i.e the model goes through each interview chunk separately, and thus can duplicate codes), I used a previous strategy of reducing each codebook (see De Paoli, 2023), by merging very similar codes. After the reduction, the codebook was n=39 for the challenges and n=75 needs. Examples of codes generated with this prompt can be seen in Tables 5-8, in the section 'Evaluation' of the paper.

**Finding Themes (phase 3)**

The subsequent phase of TA (following Braun & Clarke process) entails sorting and grouping the codes (from phase 2) in themes. In essence codes are features in the data that the analyst identifies, and themes are patterns in the data which encapsulate multiple codes. For this purpose, I used the prompt shown in Figure 3. This is very similar to the one I used in my previous work (again see De Paoli, 2023). In this case I asked the LLM (for both the list of code-challenges and code-needs, separately) to generate 12 groups (or themes). This number (12) is, to an extent, arbitrary, however I estimated it as follows: the EUREKA project had produced 6 personas from the dataset of farmers, I thus estimated that for building one persona I would need to use 2 theme-challenges and 2 theme-needs (so 6 *2 for each theme book).

```
prompt = f"""

Determine how all the topics in the following list of topics can be grouped together,
and topics can also be in more than one group. \
Group all the topics numbers only and provide a name and a description for each group\
Create 12 significant groups\

Explain me which topics fit in more than one group\
Display the full list

List of topics: {", ".join(topic_list)}

"""
```

**Figure 3** – Prompt used to sort codes into themes

Four full examples of themes (for challenges and needs) are presented in Tables 5-8 in the section 'Evaluation' of the paper, including the name of the theme, its description (as provided by the model) and the list codes, descriptions and related quotes. Table 1 and Table 2 instead present the full list of theme-challenges and theme-needs, with name, descriptions and the number of underlying codes, generated by the model in this phase (with the prompt from Figure 3).

| Nr. | Theme-Challenge | Description | Nr of codes |
|---|---|---|---|
| 1 | Limited Access to Information | This group contains topics related to the challenges faced by farmers in accessing information, either due to limited network availability, poorly maintained websites, language barriers, or a lack of reliable sources. | 8 |
| 2 | Trustworthiness of Online Information | This group contains topics related to the challenge of verifying the trustworthiness of online information. It includes topics such as filtering through online information, identifying misleading information, and finding reliable sources. | 7 |
| 3 | Digital Tools and Technologies | This group contains topics related to the use and challenge of digital tools and technologies, including finding effective tools, verifying their effectiveness, and ensuring user-friendliness. | 9 |
| 4 | Education and Training | This group contains topics related to education and training in agriculture, including language barriers, a lack of formal education, and the importance of learning by doing. | 3 |
| 5 | Regulations and Compliance | This group contains topics related to regulations and compliance in agriculture, including difficulties faced by farmers in calculating fodder/feed rations and managing data related to animal fertility. | 4 |
| 6 | Navigating Online Information | This group contains topics related to the challenge of navigating through vast amounts of online information to find relevant and accurate sources. | 3 |
| 7 | Innovation and Change | This group contains topics related to the use of digital tools to innovate and optimize work in the agricultural sector. | 2 |
| 8 | Business Planning | This group contains topics related to the challenge of finding information for business planning, particularly for innovative crops. | 2 |
| 9 | Personal Connection | This group contains topics related to the concern about the lack of personal connection and the superficiality of digital communication in the agricultural world. | 1 |
| 10 | Advocacy and Sharing Information | This group contains topics related to the challenge of connecting advocacy and professional organizations to share information. | 1 |
| 11 | Language Barriers | This group contains topics related to the challenge of language barriers faced by farmers. | 1 |
| 12 | Access to Agricultural Machinery | This group contains topics related to the challenge of finding reliable information on agricultural machinery. | 1 |

**Table 1**– Theme-challenges generated by the model in phase 3

| Nr. | Theme-Need | Description | Nr of codes |
|---|---|---|---|
| 1 | Access to Information | This group includes topics related to the need for quick and easy access to information, easy access to relevant and up-to-date information, efficient filtering of scientific papers and keyword searches, and a central point of access to relevant information. | 8 |
| 2 | Digital Tools and Applications | This group includes topics related to the need for digital tools and sources to have practical relevance, practical digital tools to increase productivity, digital tools to aid in agricultural tasks and processes, and affordable digital tools and equipment. | 10 |
| 3 | Professional Exchange and Networking | This group includes topics related to the need for personal and professional exchange with colleagues and representatives, a platform to share experiences and knowledge, and connection and community within the agricultural industry. | 4 |
| 4 | Animal Health and Farming Issues | This group includes topics related to finding solutions to animal health and farming issues, optimizing work to prevent waste in the company, and monitoring and tracking poultry health and mortality rates. | 5 |
| 5 | Expert Advice and Consultation | This group includes topics related to the need for expert advice and consultation for problem-solving and informed decision-making. | 4 |
| 6 | Language and Communication | This group includes topics related to the need for communication and gathering information in a language that is understood, particularly when it comes to technical terms related to agriculture. | 1 |
| 7 | Beekeeping | This group includes topics related to the need for current and updated information on the beekeeping industry and policies, efficient data management tools for beekeeping, and digital tools to make beekeeping easier and more efficient. | 4 |
| 8 | Organic Agriculture and Ancient Seed Varieties | This group includes topics related to staying updated with information related to organic agriculture and ancient seed varieties. | 1 |
| 9 | Mapping and Documentation | This group includes topics related to the need to use digital applications for mapping and documentation purposes. | 1 |
| 10 | Learning and Education | This group includes topics related to flexible learning options, remote exams, and expert consultancy and advice. | 2 |
| 11 | Grain Growing and Livestock Feed | This group includes topics related to practical information on grain growing, direct sowing, and grain as feed for livestock. | 1 |
| 12 | Agricultural Machinery and Equipment | This group includes topics related to reliable information sources on agricultural machinery and equipment, objective comparison or testing of agricultural products, and finding sensors specific to agriculture. | 3 |

**Table 2** – Theme-needs generated by the model in phase 3

**Evaluating Themes (phase 4)**

In this phase of TA, the analyst re-evaluates the themes, identifies which ones are indeed solid themes and which may just be sub-themes or only codes. This is a phase of re-evaluation of the previous phase and solidification of the TA. In De Paoli (2023), I suggested using the parameter Temperature (T) of the model for this evaluation. This parameter (alongside a similar parameter top_p) influences the response of the model with a degree of randomness or 'creativity'. The parameter accepts values between 0 (no randomness) and 1 (max randomness). All the previous analysis (phase 2 and 3) was done with T at 0 (zero). The approach I experimented with for phase 4, thus far, is to change the T parameter to a higher value to see if some themes appear consistent and if others emerge. This time I also introduced a further aspect for the potential review: the number of codes of each theme as presented in phase 3. Indeed, a few themes generated in phase 3, are composed of just one code (see Tables 1 and 2). This may signal that the theme could be weak and just a code (rather than a theme), and that operating the model with a higher T at phase 4 may tell us new candidates' themes. Four theme-challenges (Nr. 9, 10, 11, 12 - Table 1) and the same number of theme-needs (Nr. 6, 8, 9, 11 - Table 2) are only composed of one code. We also need to remember that the list of code-challenges is smaller than that of code-needs, and therefore the model had a small set of codes to build theme-challenges. Moreover, we need to remember that the themes were built out of a reduction of the codebook, therefore even if they only have one code, in fact this code might have appeared in multiple interviews.

Table 3 shows three sets of themes generated at T=0.5 for the theme-challenges (and the associated number of codes). Table 4 shows the same for the theme-needs. We are looking in these tests for consistency, but potentially also for new themes replacing previous themes with just one code.

| Nr. | Theme (Test_1) | Nr Codes | Theme (Test_2) | Nr Codes | Theme (Test_3) | Nr Codes |
|---|---|---|---|---|---|---|
| 1 | Limited Access to Information | 10 | Limited Access to Information | 12 | Limited Access to Information | 10 |
| 2 | Trustworthiness of Online Sources | 3 | Trustworthiness of Online Sources | 3 | Trustworthiness of Online Sources | 3 |
| 3 | Filtering through Online Information | 2 | Digital Tools and Innovations | 2 | Filtering Information | 2 |
| 4 | Language Barriers | 1 | User-Friendliness of Digital Tools | 1 | Language Barriers | 1 |
| 5 | Digital Tools for Agriculture | 5 | Crop and Animal Management | 1 | Technical Assistance | 1 |
| 6 | Education and Training | 3 | Education and Training | 2 | Education and Training | 2 |
| 7 | User-Friendliness of Digital Tools | 1 | Business Planning and Differentiation | 5 | Digital Tools | 5 |
| 8 | Business Management | 4 | Resistance to Change | 1 | User-Friendliness | 1 |
| 9 | Forest Management | 2 | Advocacy and Professional Organizations | 1 | Resistance to Change | 1 |
| 10 | Advocacy and Professional Organizations | 1 | Personalization of Platforms | 1 | Small Market | 1 |
| 11 | Personalization of Digital Tools | 1 | Reliability of Digital Sources | 3 | Reliability | 3 |
| 12 | Superficiality of Digital Communication | 1 | Superficiality of Digital Communication | 1 | Search Results | 3 |

**Table 3** – List of theme-challenges generated with T at 0.5 (3 tests) and number of codes

We can now compare Table 3 with Table 1 and look for consistency of (and thus solid) themes and for potential candidates' themes to replace those with only one code. What we can observe is that the themes in Table 1, from Nr 1 to 9 appear consistently also in Table 3. We can preliminary assume these are all valid themes to keep. The code Nr. 11 in Table 1 ('Language Barrier'), despite having only one code, is consistent across Table 3 and should probably be kept. In fact, if we look at Table 6 in the 'Evaluation' section of the paper (which reports the full challenge-theme 'Language Barriers') we can see these barriers emerged across multiple interviews. The remaining 3 themes with just one code in Table 1 ('Personal Connection', 'Advocacy and Sharing Information' and 'Access to Agricultural Machinery') do not seem to appear in Table 3 and may be candidates for clear replacement. In Table3, the theme 'Superficiality of digital communication' appears twice (also with just one code), upon reading the description of this new theme, this seemed to overlap with the theme 'Personal Connections' (or the lack thereof) from Table 1. The original theme is then kept. The theme 'Reliability' appears in both Test_2 and Test_3 in Table 3 with three codes and is a potential theme that can replace the theme on 'Agricultural Machineries', due to the description of this theme referring to reliability (which also appear in the theme 'Agricultural Machineries' from Table 1). The theme 'Advocacy' from Table 1 appears also in Test_1 in Table 3. This may therefore be a good candidate to keep (despite the single code). Thus, the new list of theme-challenges from Table 1 will see the replacement of the theme Nr. 12 ('Agricultural Machineries'), with the theme 'Reliability' (from Test_3, Nr. 11, Table 3), which is highlighted in yellow.

| Nr. | Theme (Test_1) | Nr Codes | Theme (Test_2) | Nr Codes | Theme (Test_3) | Nr Codes |
|---|---|---|---|---|---|---|
| 1 | Access to Information for Agriculture and Farming | 9 | Access to Information Platforms | 7 | Access to Information | 7 |
| 2 | Digital Tools for Agriculture | 14 | Digital Tools for Farm Management | 15 | Digital Tools and Applications | 12 |
| 3 | Personal and Professional Exchange | 4 | Expert Knowledge and Consultation | 5 | Expert Advice and Consultation | 3 |
| 4 | Language and Communication in Agriculture | 1 | Language and Communication | 1 | Animal Health and Farming Issues | 4 |
| 5 | Animal Health and Husbandry | 6 | Animal Health and Husbandry | 5 | Professional Exchange and Networking | 5 |
| 6 | Sustainability and Environmental Impact | 2 | Sustainable Agriculture | 2 | Local Information Sources | 1 |
| 7 | Digital Competence | 1 | Equipment and Machinery | 2 | Regulatory Compliance | 1 |
| 8 | Beekeeping | 3 | Knowledge Sharing and Community | 3 | Access to Trustworthy Information | 2 |
| 9 | Waste Reduction and Optimization | 3 | Personalization and User-Friendliness | 2 | Innovative Solutions | 1 |
| 10 | Community and Connection in Agriculture | 5 | Efficiency and Waste Reduction | 3 | Language and Learning | 2 |
| 11 | Expert Consultancy and Advice | 3 | Remote Learning and Support | 2 | Community and Connection | 2 |
| 12 | Personalization and User Experience | 2 | Specific Industry Knowledge | 4 | Grain Growing | 1 |

**Table 4** – List of theme-needs generated with T at 0.5 (3 tests) and number of codes

We can operate on a similar line now with the theme-needs. From Table 2, themes 1 to 7 will be kept as valid since they reappear consistently in various shapes in Table 4. Also, in this case the theme 'Language and Communication' despite having only one code, is kept due to its consistency across the three tests at T 0.5. The themes 10 and 12 from Table 2 also will be kept as they reappear in Table 4 and have more than one code. A decision needs thus to be made about the remaining themes with just one code from Table 2 (respectively 'Organic Agriculture and Ancient Seed Varieties', 'Mapping and Documentation' and 'Grain Growing and Livestock Feed'). In Table 4, unlike Table 2, we see the issue of 'Personalization' appearing twice, this is a good candidate theme, also directly related with specific needs for digital tools (we will use the Nr. 9 from Test_2 in Table 4, for our final list of themes). Also, the theme on 'Community' appears in all three tests and not in Table 2, this is also a good candidate theme (we will use the Nr. 8 from Test_2, Table 4). Lastly the theme of 'Waste Reduction' also appears in Test_1 and Test_2 but not in Table 2, hence it will be kept as theme (we will reuse Nr 10 from Test_2, Table 4). The theme-needs from phase 4 adopted for the final theme-needs book (replacing three weak themes from Table 2), are highlighted in yellow in Table 4.

The final theme books (for both challenges and needs) thus encompass the changes to the themes as presented above, with one change for the theme-challenges, and three changes for the theme-needs. Moreover, the final theme books include all the previous codes, descriptions and quotes (identified in phase 2, before the reduction of the codebook). Again, for examples see Tables 5-8 in the section 'Evaluation' of the paper.

The process of conducting a TA also entails a phase 5 related with renaming and summarising themes. As said earlier, this phase is not performed here, and readers can consult De Paoli (2023) for an approach to operate that phase. Will focus now on using the two theme books (challenges and needs) to conduct phase 6 of a TA: the writing of the results.

## Writing up user personas

We have a list of 12 theme-needs and 12 challenges-needs from the TA. We will attempt to build user personas with this analysis. In phase 3 (when estimating the number of themes to be inferred) I had assumed that one persona would be built using 2 theme-challenges and 2 theme-needs. Therefore, since not all the themes will be used (and are not normally used by human writers of personas), we need to operate a selection. There may be different ways of deciding which themes should compose a single persona. For example, we could have the human analyst decide which theme-needs and theme-challenges to use, or we can use a randomised process of selection of the themes, or a mix of the above two. For this simple experiment, I wrote python code to select at random two sets of themes to be used (i.e 2 theme-challenges and 2 theme-needs), as described in Figure 4. This was done by creating a list of random tuples for each theme book (6 tuples of 2 themes each) and then selecting one tuple. Every time that the script is run, the list of tuples is randomised again. This randomised process entails in fact 78 combinations for each theme book (thus potentially allowing 6084 combinations across all themes). The selected theme-challenges and theme-needs are then passed to the model via the prompt, to generate a user persona. To increase the variability of the personas I also operated the model with Temperature set a 1, but this is just a choice to have some more 'creativity' from the model in its response.

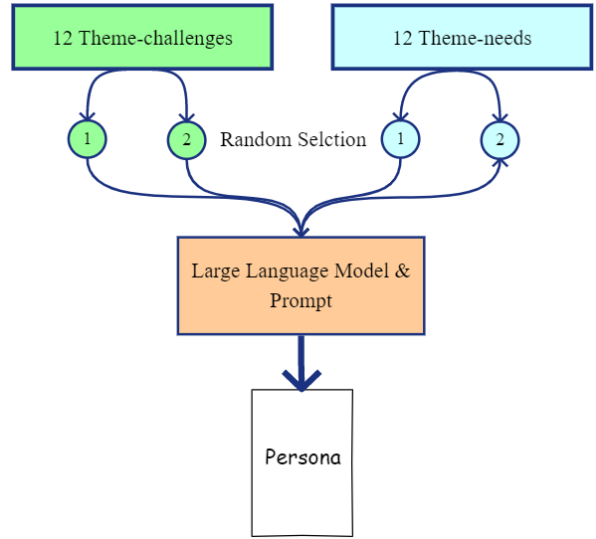

**Figure 4** – Themes (tuples) selection and personas writing

Note also that when building personas in UCD, normally there are more dimensions included such as attitudes, behaviour, interests and so on. Here we are working only on two theme-dimensions (as we are attempting to establish a process), however it would be possible to have more themes related to other dimensions of personas. It is sufficient to reproduce the previous steps of the analysis to generate these additional themes. However, several theme-dimensions may impact the limit of the model in relation to the max tokens.

Figure 5 presents the prompt used for generating personas. This deserves some comments.

```
prompt = f"""

Using the needs and challneges in the lists provided below, write a user persona, from a specific EU country\
Based on the needs tell also if the persona is young, middle age or old and what is the persona main goal (max 1)\
include also a narrative background (max 200 words)

Also identify with bullet points the main needs (max 3, max 30 words each)\
and main challenges (max 2, max 20 words each) of the persona.
Moreover tell me the persona IT skills (low, medium, high),\
and the persona attitude toward digital innovation (low, medium, high).

Lastly also include a quote taken from either list (of needs or challenges) representing the goal

List of needs: {", ".join(needs_list)}
List of challenges: {", ".join(challenges_list)}

"""
```

**Figure 5**- Prompt used for the generation of a use persona

First the randomly selected tuples of themes (2 theme-needs and 2 challenge-needs) are passed to the model as lists ('List of needs', 'List of challenges'), each element of a tuple includes the theme name, the theme description (from phase 3-4) and all the underlying codes and the associated quotes (phase 2-4). This is the material (i.e. the analysis results) that we ask the model to use for building the persona. In the prompt, we ask the model to include the persona's: goal, background description, a vague indication of the age, three needs, two challenges, and a quote illustrating the goal and whether the persona is tech savvy and open to innovation.

An important aspect of this prompt is that it is not telling the model anything about what kind of personas to build, like who the persona is (what the persona does, her job, her goal, her motives etc). It just informs the model of what formal element to include (e.g. a goal, a quote, three needs). This is significantly different from the examples of prompts we can find online related to asking chatGPT to generate user personas, where details about the persona need to be provided to the model. For instance, in the literature review we mentioned one such prompt (taken from a youtube video tutorial), where it was specified that the persona to be generated was a busy mum, living in a city, with kids, concerned about health food and career. Another example taken from an online tutorial (this time of an audience persona) is "Create an audience persona for a business that sells trendy fashion wear for young adults, Colorful clothing and jewelry. The business sells at reasonable prices so all young adults can purchase from this brand"[7]. This level of details is not needed in the prompting when the persona is generated using results of a TA, and the prompt only needs to provide indication about the formal structure or components of the persona narrative. I only added one detail to the prompt that the persona should be from a 'from a specific EU country'. Indeed, the prompt (without this specific instruction) would sometimes provide also the country in which the persona works but sometimes it would not, and this I believe is dependent upon countries being named or not in the data that the model uses (i.e. tuples).

What follows are two examples of personas generated with the above prompt. The persona Katharina was generated using (with the random selection) the following tuple of theme-challenges: (1) Navigating Online Information and (2) Language Barriers; and the following tuple of theme-needs: (1) Personalization and User-Friendliness and (2) Animal Health and Farming Issues. The persona Gisela was generated using the following tuple of theme-challenges: (1) Language Barriers and (2) Personal Connection; and the following tuple of theme-needs: (1) Agricultural Machinery and Equipment and (2) Knowledge Sharing and Community.

I structured the personas in a table and included a picture which I generated with Dall-e (https://openai.com/product/dall-e-2), mostly to attempt to reproduce some of the empathy that user personas seek to achieve. However, what we are interested in considering is the textual narrative and whether the textual description of these personas do appear credible (i.e. 'fictional, yet realistic' - Harley, 2015). This will be discussed in the next section of the paper. Please note that in the appendix there are some further examples of personas generated from the analysis, using the prompt of Figure 4. Also note the order of items of the personas are presented in the way they were produced by the model.

---

[7] See https://www.youtube.com/watch?v=db3zlVeAMhE

| 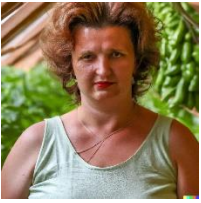 | **Name:** Katarina<br>**Age:** Middle-aged<br>**Country:** Poland |
|---|---|

**Narrative Background:** Katarina is a middle-aged farmer based in Poland. She inherited her family farm and has been managing it for over a decade. She is passionate about farming and aims to optimize her productivity while following ethical farming practices. Since there are limited resources and solutions available to her in her area, Katarina has to rely mostly on online information to find solutions to her problems related to animal health and farming practices. However, she faces language barriers and struggles to maneuver through the vast amount of information available online. Katarina values user-friendly digital tools and prioritizes the need for problem-solving and optimization of her farming practices.

**Main Goal:** To find innovative and effective solutions to improve animal health and optimize her farming practices.

**Main Needs:**
- Filtering information sources to ensure the accuracy of the information
- Finding solutions to problems related to animal health and farming practices
- Optimizing work in the company and avoiding waste

**Main Challenges:**
- Difficulty accessing information due to language barriers
- Navigating through vast amounts of online information
- Filtering out misleading information from the internet

**IT Skills:** Medium
**Attitude Toward Digital Innovation:** High

**Quote Representative of the Goal:** "My opinion is that digital sources can help optimize work on the farm. If you have tools that allow you to work better, the entrepreneur works better, the employees and the company work better, and so on. Maybe you can also avoid a lot of waste."

| 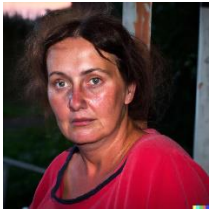 | **Name:** Gisela Schmidt<br>**Age:** Middle-aged (45)<br>**Country:** Germany |
|---|---|

**Main Goal:** To find reliable information sources on agricultural machinery and equipment specific to agriculture.

**Background:** Gisela Schmidt is a middle-aged farmer from Germany who runs a small farm with her family that specializes in crop production. She has been in the agricultural industry for over 20 years and has seen its evolution with the introduction of technology and digitalization. While Gisela has a basic understanding of digital tools, she faces challenges in finding reliable information and sources specific to agriculture. Her main concern is finding reliable information sources on agricultural machinery and equipment, including sensors specific to agriculture. Gisela believes that she can improve crop production and save time and money if she can find the right sensors and equipment. She often spends a lot of time searching for information online but struggles to find trustworthy sources.

**Main Needs:**
- Reliable information sources on agricultural machinery and equipment
- Finding sensors specific to agriculture
- Objective comparison or testing of agricultural products

**Main Challenges:**
- Language and education barriers
- Difficulty accessing reliable information
- Lack of personal connection and superficiality of digital communication

**IT Skills:** Medium
**Attitude toward digital innovation:** Medium

**Quote:** "Then I used the keywords of those research papers to find other producers, not the salesmen, of the sensors but producers. And then I got in contact with the dual M guys, but also with Medusa, with all kinds of software and hardware and software suppliers that are building nice sensors, but not specifically for agriculture."

## Evaluation

Overall, the background of each of the two examples of persona (Katharina, Gisela) offers important clues about their story, such as the kind of farm they operate, what the farm produces (e.g. crops), what are the personas intentions (e.g. efficiency, saving time with sensors). It tells also something about the relation of the persona to the farm (a farm inherited from family or a farm run with the family), or about the issues related with keeping up with digital technologies in the sector and online information. To evaluate the validity of these narratives we will look at how the underlying themes appear and are composing the personas. We will do so for the persona Katharina.

Table 5 and 6 show the theme-challenges used for building 'Katharina' and Table 7 and 8 the theme-needs.

| Challenge | Navigating Online Information | |
|---|---|---|
| Description | This group contains topics related to the challenge of navigating through vast amounts of online information to find relevant and accurate sources. | |
| Codes | Description | Quote |
| Filtering Information Online | The interviewee faces the challenge of filtering through the vast amount of information available online to find relevant and useful information. They have developed a strategy of using specific keywords and ruling out advertisements to find what they need. | Usually I find a specific keyword and then add another keyword to that and add another keyword to that. But I don't type in all the keywords at once. If you do it step by step, you get more information for every goal and you easily get the rubbish out. |
| Filtering Information Online | The interviewee faces the challenge of filtering information sources to ensure the accuracy of the information. | So I'm looking informations. And when I get many results, I'm looking at who got this information. How did they get there? If it's still trials, if it's a trial site or if it's only guesses and maybe just a reflection of it could be like that. |
| Filtering Information Online | The interviewee faces the challenge of filtering out misleading information from the internet, which can be both digital and non-digital. | Misleading information can be digital or non-digital in this there is no difference. Because there is more digital information, there will be more misleading ones. |
| Navigating Information | The interviewee struggles with navigating through the vast amount of information available online to find reliable sources. | Reliability seems at first glance, and also when someone talks about something they've already tried. Once you have the routine, you can see what advertising is. Good infographics help with comprehension, but they are worth nothing if they are not grounded, they are just advertising. |
| Information Overload | The challenge of sifting through a large amount of search results to find the right information. | Of course, if you go, for example, to Google and you search something, you will have a lot of outputs. So, it's very difficult to spot the right ones, so sometimes it will take a lot of time searching. |

**Table 5** – Theme-challenge 'Navigating Online Information

| Challenge | Language Barriers |
|---|---|
| Description | This group contains topics related to the challenge of language barriers faced by farmers. |

| Codes | Description | Quote |
|---|---|---|
| Language and Education Barrier | Difficulty accessing information due to language and education barriers | The content must be available in the respective national languages and also for users with different levels of education - not in all countries is agricultural education as good as here in Germany. |
| Language and Education Barrier | The challenge of language barriers for farmers who may not be proficient in foreign languages and need access to information in their local language. | Of course a farmer is also a firm manager and he's has his own company and he needs to to manage his company. So he needs to know about numbers, about accountability, needs to know about politics. But it shouldn't take him out off core business, which is agronomy. So as a tool, must must help the farmer and the user in that direction. So learning a different language, I don't think it's it's the best way to try to be helpful. |
| Language and Education Barrier | The interviewee faces a challenge of accessing information in languages other than Slovakian or Hungarian. | We mostly use Slovak or Hungarian, not everyone understands English. Perhaps the main information can be understood in German. Access to information is limited to these languages. |
| Language and Education Barrier | Difficulty in understanding technical terms in English while searching for cultivation problems online | If I have to go and look for cultivation problems, in English there is no way I can do it. |
| Language and Education Barrier | The interviewee only looks for information in Polish, limiting access to potentially valuable information in other languages. | On our websites we have a lot of information but time to time we are checking abroad website and translating it because some websites are better than polish. For example some Belgium webpages have better information than polish. |

**Table 6** – Theme-challenge 'Language Barriers'

| Need | Personalization and User-Friendliness |
|---|---|
| Description | This group includes topics related to the need for a certain degree of personalization based on what the user does or would like to have access to in terms of information. It also includes the need for digital tools that are easy to use and accessible. |

| Codes | Description | Quote |
|---|---|---|
| Personalization | The need for a certain degree of personalization based on what the user does or would like to have access to in terms of information. | Yes. But I also immediately understand why this isn't possible for everyone. Some people may not know how to configure such an app themselves. |
| User-Friendly Digital Tools | The need for digital tools that are easy to use and accessible | It should be user friendly...They don't have good digital skills. And so everything should be very |

| | | intuitive. Yes, very easy tool to use. This is the main argument. |

**Table 7** – Theme-need 'Personalization and User-Friendliness'

| Need | Animal Health and Farming Issues | |
|---|---|---|
| **Description** | This group includes topics related to finding solutions to animal health and farming issues, optimizing work to prevent waste in the company, and monitoring and tracking poultry health and mortality rates. | |
| **Codes** | **Description** | **Quote** |
| Information Discovery | The need to discover new information and techniques related to farming and animal husbandry. | Steal with your eyes to learn how someone else does it, and what you can do better yourself. |
| Problem Solving | The need to find solutions to problems related to animal health and farming practices. | For example, let's take the (lack of) fertility of the animals. We talked about it with our vet. He drew blood and sent a blood sample to the lab. They examined it and found that the cows had a deficit of a certain mineral. |
| Problem Solving | The need to effectively solve problems using digital tools and platforms. | I do had a situation when the field was... |
| Problem Solving | The need to optimize work in the company and avoid waste | My opinion is that digital sources can help optimize work in the company. If you have tools that allows you to work better, the entrepreneur works better, the employees and the company works better and so on. Maybe you also avoid a lot of waste. |
| Fauna Management | The need to manage fauna in a way that is effective and humane | Roe Deer? They seem strange things. How to drive away the Roe Deer because objectively there are no resources for you... they only sell you products. But maybe, for example, this is something that if there were answers in Europe on how to manage the fauna, it would be interesting because everyone have an experience. |
| Optimization | The need to optimize work in the company and avoid waste | My opinion is that digital sources can help optimize work in the company. If you have tools that allows you to work better, the entrepreneur works better, the employees and the company works better and so on. Maybe you also avoid a lot of waste. |
| Monitoring and Tracking | The need to monitor and track poultry health and mortality rates to identify potential problems and make informed decisions. | For mortality and temperature, I made a Google Sheets. I monitor these and then I make the charts, I make them myself. |

**Table 8** – Theme-need 'Animal Health and Farming Issues'

If we look back at Katharina, her goal and the quote come from the code 'optimization', in the theme-need 'Animal Health and Farming Issues' (highlighted in light yellow, in Table 8). Information in the background narrative of Katharina mentions aspects such as lack of language skills and the need for user friendly tools which reflect both one of the challenges (Table 6) and one of the needs (Table 7). Specific main needs and challenges of Katharina are also identifiable in the tables. For example, Katharina does seem to have a strong focus on the navigation and identification of information. One of her challenges, relates to filtering out misleading information from the Internet, is clearly reflected from one of the codes in Table 5 (highlighted in green). One of her needs, related to finding information about animal health, is clearly reflected in one of the 'problem solving' codes in Table 8 (highlighted in green). Language barriers also is another challenge that she faces, and although all the codes in Table 7 may have contributed to this, it is likely that that model used the one related to information in Polish (highlighted in green).

## Discussion

The main goal of this paper was to conduct a TA with an LLM (GPT3.5-Turbo), covering also phase 6 of the approach proposed by Braun & Clarke (2006), the writing of the results. In practical terms the goal was to use a TA of semi-structured interviews to create user personas. In this discussion we will reflect on the results of this research.

First, this research has been an opportunity to test again the embryonal and crude process for conducting an inductive TA with an LLM, I proposed in my previous paper (De Paoli, 2023). Overall, this process can be reproduced on different datasets and it is clear that it can deliver some meaningful (but still basic) codes and themes. It certainly will need more work and refinement, but I consider it as a good initial basis for reflection on how to conduct a TA with an LLM. In this paper I also covered in more detail phase 4, where an evaluation of themes has been done and weak themes have been replaced with other themes which appeared more valid and sound for the analysis. This clearly signals the fact that a TA with an LLM is done within the frame of a Human-AI collaboration.

Second, using the themes produced from the process of analysis, it does seem that the LLM can indeed write up the results, in this case in the form of user personas. With a formal prompt (which does not give any indication about the content, except the structure) and a set of themes, the model can write up a persona narrative. We have also seen (in one example) how specific codes are reflected in such a narrative, including in the persona's goal (and related quote), background or in the persona needs and challenges. At this stage, it is difficult to tell in what ways the model prioritises certain codes over others for building the narrative. However, this is a data driven process of writing up personas, and the personas are a reasonable reflection of the data analysis, they are 'fictional, yet realistic' (Harley, 2015). I consider therefore, that we can provide (at least initially) a positive answer to the problem of covering phase 6 of a TA, as indeed also this phase can be attempted with an LLM.

Some words need also to be used to reflect on personas, the field of UCD and the use of LLMs. The process I proposed here clearly is significantly different from the generation of entirely fictional personas, which are not based on empirical data, which we often see in online tutorials. Those entirely fictional personas clearly cannot be a good ('yet realistic') representation of target users, as they are not derived from user research.

I would also argue that the personas generated by the LLM using a TA of semi-structured interviews, are probably to be considered not finished personas, but more like initial prototypes, which would require further refinement by the human analysts. Indeed, the model can generate multiple ideas for the personas, by mixing themes (in my example I mixed 2 theme-needs and 2 theme-challenges, and this would allow more than 6000 combinations), and then the analyst can decide which personas do seem more representative of the target user group, and eventually also enrich them with further details, in line with the proposed design work. In essence, I see the process more as an idea generation of personas, rather than as a replacement for the work of the user researcher. The important aspect is that these ideas are generated from real data and from a structured qualitative analysis and are not just entirely fictional.

**Recommendations**

In this section I offer some recommendations based on the results of this work and connect, where relevant, some of these recommendations with previous literature.

**Human-AI Collaboration.** The conduction of a qualitative analysis with an LLM is a process of collaboration between the model and the human analysts. This confirms the previous observations by Gao et al (2023) and Jiang et al. (2021). However, we need to link the Human-AI collaboration better to what could be the methodological implications for social sciences. Especially for the reporting of the methods underpinning research. It is clearly not enough to state that an analysis was done with an LLM, and all the steps taken by the model and by the human analyst in collaboration will also need to be reported and documented in any methodology to ensure that validity of the results.

**Prompting.** Prompting is clearly the key element of using LLMs successfully (Xiao et al., 2023) and the reporting of prompts will need to become a standard practice in methods section of social sciences, when (and if) the analysis is done with an LLM. Moreover, practices for improving prompts will need to be shared across the social sciences research community to improve the capacity to conduct qualitative analysis, such as TA or the production of intermediate writing up of the results.

**Coding.** In phase 4 of the TA proposed in the previous pages, I have done an evaluation of the themes and replaced weak themes with stronger ones. A similar evaluation, however, should also be conducted in the codes generated at phase 2. Upon scrolling the list of codes, for example I found one code with a truncated quote (this can be seen in Table 8, in one of the 'Problem Solving' codes) and one code with a quote that was not very representative. There is in other words, room for improving the initial coding, and it may be possible to envision also a process which relies on the randomness of the model for doing this. However, this operation is clearly costly as it requires multiple processing of a dataset with the model.

**Publishing.** There is debate about the use of LLMs in academic publishing (see e.g. Thorp, 2023; Lund et al., 2023). Whilst the use of these models in place of the human author (or for manipulating results) should be rejected, it may be acceptable to use text written in a Human-AI collaboration, in the form of intermediate products. For example, if user personas generated with the previous process (and then enriched by the human analyst) are included in a UCD publication (and where the process is fully detailed at methodological level), it may be acceptable for these types of intermediate textual products to be part of a manuscript. However, entering in this debate is beyond the scope of my work.

**Chat VS API.** The few existing online tutorials and few other research contributions focusing on doing qualitative/TA analysis with an LLM, use the web chat version of the model, which requires manually copy-pasting content from a textual file into the chat and then copying back the response into other documents, such as another text file. This procedure is not functional for the conduction of e.g a TA, since it does not allow further data manipulation easily or to reuse the results of the analysis at scale. For example, a TA done with the web chat, would end up in a e.g. text processor document, but this analysis cannot be reused easily for the model to write up results, such as user personas, at scale.

**API Vs Integration.** While LLMs are starting to be integrated in Qualitative Data Analysis Software[8] (which will limit probably the use of the web chat in the future), it remains unclear if these integrations will offer transparency on what is done by the model. The use of the API (although requiring some basic programming skills), offers, at least for now, a reasonable level of transparency, since all the steps conducted for an analysis can be traced, and the outputs can be reconstructed. This is important for methodological reasons and reporting.

**Limits of this work**

**Potential Bias.** For some reason, the large majority of the personas generated with the prompt (Figure 4), resulted to be middle-aged. I am not sure the reason for this, but it may be that the model has a bias and assumes people working in farming are all middle-aged. Therefore, the prompting I proposed may need to be adjusted to specify if the persona needs to be young, middle-age or else. Moreover, many of the personas appeared to be called Maria, and coming from Italy. In part this may be a bias of the model (which I cannot prove) and in part this may be a reflection on the data seen by the model.

**Two sets of themes.** I built the personas deriving them from two set of themes (needs and challenges). Clearly personas in UCD have more dimensions than two (e.g. behaviour, interests). However, my focus was only demonstrative, to see if we could write up results of a TA in a way that is at least satisfactorily. For writing more articulated personas (with more dimensions), more themes will need to be produced from the analysis.

**Phase 1**. This phase of a TA would require the familiarisation with the data. At present this cannot be done by the LLM due to the limitations in memory and tokens. We will have to assess in the future if more powerful models can perform also this phase in the future.

# Conclusion

This paper has shown that it is possible to perform phase 6 (writing up) of a Thematic Analysis with an LLM, based on the results of a TA conducted on semi-structured interviews. A dataset of 14 user interviews was analysed with an LLM to derive a set of needs and challenges as themes. These themes were then used in a prompt to the model for building user personas. Two examples of personas were presented, and one was evaluated against the underpinning themes. These personas, I would say, appear 'fictional, yet realistic'. While the process for conducting a TA with an LLM is still crude, there clearly is an initial basis for further work.

---
[8] see se.g. https://atlasti.com/

# References

Braun, V. & Clarke, V. (2006). Using thematic analysis in psychology. *Qualitative Research in Psychology*. 3 (2), 77–101. https://doi.org/10.1191/1478088706qp063oa

Clarke, V., & Braun, V. (2013). Teaching thematic analysis: Overcoming challenges and developing strategies for effective learning. *The psychologist*, *26*(2), 120-123.

De Paoli S. (2023). Can Large Language Models emulate an inductive Thematic Analysis of semi-structured interviews? An exploration and provocation on the limits of the approach and the model. Available from: https://arxiv.org/abs/2305.13014

King, M. R., & ChatGPT. (2023). A conversation on artificial intelligence, chatbots, and plagiarism in higher education. *Cellular and Molecular Bioengineering*, 16(1), 1-2. https://doi.org/10.1007/s12195-022-00754-8

Kocaballi, A. B. (2023). Conversational ai-powered design: Chatgpt as designer, user, and product. Available from: https://arxiv.org/abs/2302.07406

Floridi, L. (2023). AI as Agency without Intelligence: On ChatGPT, large language models, and other generative models. *Philosophy & Technology*, *36*(1), 15. https://doi.org/10.1007/s13347-023-00621-y

Gao, J., Guo, Y., Lim, G., Zhan, T., Zhang, Z., Li, T. J. J., & Perrault, S. T. (2023). CollabCoder: A GPT-Powered Workflow for Collaborative Qualitative Analysis. *arXiv preprint arXiv:2304.07366*. https://doi.org/10.48550/arXiv.2304.07366

Harley, A. (2015). Personas make users memorable for product team members. *Nielsen Norman Group*, *16*. Available from: https://www.nngroup.com/articles/persona/

Jiang, J. A., Wade, K., Fiesler, C., & Brubaker, J. R. (2021). Supporting serendipity: Opportunities and challenges for Human-AI Collaboration in qualitative analysis. *Proceedings of the ACM on Human-Computer Interaction*, *5*(CSCW1), 1-23. https://doi.org/10.1145/3449168

Laubheimer, P. (2020). Persona Types: Lightweight, Qualitative, and Statistical. *Nielsen Norman Group*, *21*(06), 2020. https://www.nngroup.com/articles/persona-scope/

Lund, B. D., Wang, T., Mannuru, N. R., Nie, B., Shimray, S., & Wang, Z. (2023). ChatGPT and a new academic reality: Artificial Intelligence-written research papers and the ethics of the large language models in scholarly publishing. *Journal of the Association for Information Science and Technology*. 74(5), 570-581 https://doi.org/10.1002/asi.24750

Miaskiewicz, T., & Kozar, K. A. (2011). Personas and user-centered design: How can personas benefit product design processes?. *Design studies*, *32*(5), 417-430. https://doi.org/10.1016/j.destud.2011.03.003

Nielsen, L., & Storgaard Hansen, K. (2014, April). Personas is applicable: a study on the use of personas in Denmark. In *Proceedings of the SIGCHI Conference on Human Factors in Computing Systems* (pp. 1665-1674).

Matthews, T., Judge, T., & Whittaker, S. (2012, May). How do designers and user experience professionals actually perceive and use personas?. In *Proceedings of the SIGCHI conference on human factors in computing systems* (pp. 1219-1228).

O'Connor, S. (2022). Open artificial intelligence platforms in nursing education: Tools for academic progress or abuse?. *Nurse Education in Practice*, *66*, 103537-103537.

Rosala, M. (2019). How to analyze qualitative data from UX research: Thematic analysis. *NN-Nielsen Norman Group*. Available from: https://www.nngroup.com/articles/thematic-analysis/

Salazar, K. (2020). Just-right personas: How to choose the scope of your personas. *Nielsen Norman Group. Retrieved*, *23*, 2020. Available from: https://www.nngroup.com/articles/persona-scope/

Siemon, D. (2022). Elaborating team roles for artificial intelligence-based teammates in human-AI collaboration. *Group Decision and Negotiation*, *31*(5), 871-912. https://doi.org/10.1007/s10726-022-09792-z

Thorp, H. H. (2023). ChatGPT is fun, but not an author. *Science*, *379*(6630), 313-313. https://doi.org/10.1126/science.adg7879

Turner, A. M., Reeder, B., & Ramey, J. (2013). Scenarios, personas and user stories: User-centered evidence-based design representations of communicable disease investigations. *Journal of biomedical informatics*, *46*(4), 575-584. https://doi.org/10.1016/j.jbi.2013.04.006

Szabolcs Vágó, & Pieter Spanoghe. (2023). EUREKA WP2: Regional Interview Transcripts [Data set]. Zenodo. https://doi.org/10.5281/zenodo.7571028

UK Government (2017). *Learning about users and their needs.* Available from: https://www.gov.uk/service-manual/user-research/start-by-learning-user-needs#:~:text='User%20needs'%20are%20the%20needs,so%20achieve%20their%20policy%20intent

Van der Cruyssen L. (2021). D2.3. Report on end-user archetypes, end-user journeys and validation workshops. Available from https://eureknos-eureka.fra1.digitaloceanspaces.com/production/deliverables/D2.3_Enduser_archetypes_journeys_WP2.pdf

Vössing, M., Kühl, N., Lind, M., & Satzger, G. (2022). Designing transparency for effective human-AI collaboration. *Information Systems Frontiers*, *24*(3), 877-895. https://doi.org/10.1007/s10796-022-10284-3

Xiao, Z., Yuan, X., Liao, Q. V., Abdelghani, R., & Oudeyer, P. Y. (2023, March). Supporting Qualitative Analysis with Large Language Models: Combining Codebook with GPT-3 for Deductive Coding. In *Companion Proceedings of the 28th International Conference on Intelligent User Interfaces* (pp. 75-78).

**Appendix – other examples of personas generated by the model based on the TA**

User Persona:

Name: Giuseppe Rossi
Age: Middle-aged
Goal: To gather reliable information and connect with colleagues in the agricultural industry to optimize work in his company and avoid always cultivating the same crops.

Background: Giuseppe Rossi is a farmer from Italy who has been in the agricultural industry for over 25 years. He owns a medium-sized farm and has a team of employees who work for him. Giuseppe is interested in learning more about the latest developments in agriculture and how digital tools can help optimize his work. However, he faces challenges in accessing reliable information due to language barriers and limited network availability. He also struggles with finding a central platform for accessing information and knowledge related to agriculture. Giuseppe values personal and professional exchange with colleagues and representatives in order to gain trustworthy information and community connection within the agricultural industry.

Main Needs:
 - Need to communicate and gather information in a language that is understood
 - Need for personal and professional exchange with colleagues and representatives in order to gain trustworthy information
 - Need for a platform to share knowledge and experiences in agriculture

Main Challenges:
 - Language barriers for accessing information in foreign languages
 - Limited network availability and poorly maintained websites
 - Finding reliable sources of information in the digital age

IT Skills: Medium
Attitude toward digital innovation: Medium

Quote: "Digital sources can help optimize work in the company. If you have tools that allow you to work better, the entrepreneur works better, the employees and the company work better and so on. Maybe you also avoid a lot of waste."

Persona: Anna, from Poland
Age: Middle age (between 40-60 years old)
Goal: To increase productivity and efficiency on her family-owned farm

Background: Anna is a middle-aged farmer from a rural town in Poland. She works on a family-owned farm that produces a variety of crops and livestock. With the increasing use of technology in agriculture, Anna has recognized the need to incorporate digital tools into her farming practices to increase productivity and efficiency. She has some basic computer skills, but would like to learn more about how digital tools can be implemented on her farm. Anna is also fluent in Polish but struggles to find information in English or other languages.

Needs:
- Access to affordable digital tools and equipment
- Practical and efficient digital tools that aid in agricultural tasks and processes
- User-friendly and easily accessible digital tools that provide efficient information retrieval

Challenges:
- Language barriers when accessing information in foreign languages
- Limited access to reliable sources of information
- Difficulty in understanding technical terms in English when searching for solutions online

IT Skills: Low to medium
Attitude toward digital innovation: High

Quote representing the goal: "I want to incorporate digital tools on my farm to increase efficiency and productivity, and to keep up with the changing agricultural industry."

---

User Persona: Maria Rossi

Age: Middle-aged

Goal: Maria's main goal is to have access to trustworthy information and knowledge sharing platforms to improve her farming practices.

Background: Maria Rossi is a small-scale farmer from Italy who has been working in agriculture for over 20 years. She has a small farm where she grows vegetables and raises cattle. Maria is looking for a platform where she can connect with other farmers and share her experiences to gain insight into how she can improve her crop yield and livestock production. She is tired of the limited sources of information available in her community and wants to learn from other farmers who are facing similar challenges.

Main Needs:
 - Personal and professional exchange with colleagues and representatives in the agricultural industry.
 - Access to a platform for knowledge sharing in agriculture.
 - Connection and community within the agricultural industry.

Main Challenges:
 - Navigating through the vast amount of online information available to find reliable sources and avoid misleading information.
 - Identifying crop issues using digital tools and validating them in the field.
 - Limited availability of economically efficient digital tools.

IT Skills: Medium

Attitude toward Digital Innovation: Medium

Quote: "I would like to have much more information in one platform. I use a lot of Facebook. I use Facebook more for work than to socialize. It is an easy way to follow the news from all the Italian newspapers and also the international newspapers."

---

User Persona:

Name: Marta
Age: Late 30s
Country: Italy
Background: Marta inherited her family's farm, located in a small village in Southern Italy. Despite having a degree in agriculture, she struggled to manage the farm and keep it up to date with the latest technologies. She is now motivated to bring her farm back to its prime and generate more revenue, but she has limited knowledge of digital tools and faces language barriers while searching for technical information online.

Goal: To access personalized and user-friendly digital tools that can help her manage her farm efficiently and access technical information in Italian.

Main Needs:
- Personalized information
- User-friendly digital tools
- Information in Italian

Main Challenges:
- Limited digital skills
- Language barriers while searching for technical information online

IT Skills: Low
Attitude towards digital innovation: Medium

Quote: "It should be user-friendly...They don't have good digital skills. And so everything should be very intuitive. Yes, a very easy tool to use. This is the main argument."